\documentclass{ifacconf}

\usepackage{shortex}
\crefname{equation}{}{} % remove Eq. prefix
\Crefname{equation}{Equation}{Equations} % 
\crefname{figure}{Fig.}{Figs.} % 
\Crefname{figure}{Figure}{Figures} % 

\usepackage{autonum}

\graphicspath{ {figures/} }
\usepackage{acro}
\acsetup{single={1},	
	uppercase/list}

\acsetup{patch/maketitle=false}
	
\DeclareAcronym{AI}{short = AI,
	long = artificial intelligence}
\DeclareAcronym{ML}{short = ML,
	long = machine learning,
	short-indefinite = an}

\DeclareAcronym{SVD}{short = SVD, long = singular value decomposition}
\DeclareAcronym{CCA}{short = CCA,
	long = canonical correlation analysis}	
\DeclareAcronym{FA}{short = FA,
	long = factorial analysis}
\DeclareAcronym{GMM}{short = GMM,
	long = Gaussian mixture model}	
\DeclareAcronym{ICA}{short = ICA,
	long = independent component analysis}	
\DeclareAcronym{LARS}{short = LARS,
	long = least-angle regression}	
\DeclareAcronym{LASSO}{short = LASSO,
	long = least absolute shrinkage and selection operator}	
\DeclareAcronym{LR}{short = LR,
	long = logistic regression}
\DeclareAcronym{PCA}{short = PCA,
	long = principal component analysis}
\DeclareAcronym{PLS}{short = PLS,
	long = partial least squares}	
\DeclareAcronym{RBC}{short = RBC,
	long = reconstruction-based contribution}

\DeclareAcronym{ANFIS}{short = ANFIS,
	long = adaptive network fuzzy inference system}
\DeclareAcronym{ANN}{short = ANN,
	long = artificial neural network,
	short-indefinite = an}	
\DeclareAcronym{BN}{short = BN,
	long = Bayesian network}	
\DeclareAcronym{CNN}{short = CNN,
	long = convolutional neural network}	
\DeclareAcronym{DNNE}{short = DNNE,
	long = decorrelated neural network ensemble}		
\DeclareAcronym{DNN}{short = DNN,
	long = deep neural network}	
\DeclareAcronym{ELM}{short = ELM,
	long = extreme learning machine}
\DeclareAcronym{GAN}{short = GAN,
	long = generative adversarial network}
\DeclareAcronym{GPR}{short = GPR,
	long = Gaussian process regression}	
\DeclareAcronym{GRNN}{short = GRNN,
	long = general regression neural network}	
\DeclareAcronym{MLP}{short = MLP,
	long = multilayer perceptron,
	short-indefinite = an}
\DeclareAcronym{RBFNN}{short = RBFNN,
	long = radial basis function neural network,
	short-indefinite = an}
\DeclareAcronym{RNN}{short = RNN,
	long = recurrent neural network,
	short-indefinite = an}
\DeclareAcronym{RT}{short = RT,
	long = regression tree,
	short-indefinite = an}			
\DeclareAcronym{RVM}{short = RVM,
	long = relevance vector machine,
	short-indefinite = an}		
\DeclareAcronym{SFA}{short = SFA,
	long = slow feature analysis}	
\DeclareAcronym{SVM}{short = SVM,
	long = support vector machine}
\DeclareAcronym{TL}{short = TL,
	long = transfer learning}	
\DeclareAcronym{VAE}{short = VAE,
	long = variational autoencoder}	
\DeclareAcronym{WNN}{short = WNN,
	long = wavelet neural network}				

\DeclareAcronym{RL}{short = RL, long = reinforcement learning, short-indefinite = an}
\DeclareAcronym{A3C}{short = A3C,
	long = asynchronous advantage actor-critic}	
\DeclareAcronym{ADP}{short = ADP,
	long = approximate dynamic programming}
\DeclareAcronym{IQC}{short = IQC, long = integral quadratic constraint}
\DeclareAcronym{BIBO}{short = BIBO, long = {bounded-input, bounded-output}}
\DeclareAcronym{SISO}{short = SISO, long = {single-input, single-output}}
\DeclareAcronym{MIMO}{short = MIMO, long = {multiple-input, multiple-output}}
\DeclareAcronym{DDPG}{short = DDPG,
	long = deep deterministic policy gradient}
\DeclareAcronym{DQN}{short = DQN,
	long = deep $Q$-network}
\DeclareAcronym{HJB}{short = HJB,
	long = {Hamilton-Jacobi-Bellman}}
\DeclareAcronym{MPC}{short = MPC, 
	long = model predictive control,
	long-plural-form = model predictive controllers,
	short-indefinite = an}		
\DeclareAcronym{PI2}{short = $\text{PI}^2$,
	long =policy improvement with path integrals}
\DeclareAcronym{PID}{short = PID, 
	long = proportional-integral-derivative}
\DeclareAcronym{PI}{short = PI, 
	long = proportional-integral}
\DeclareAcronym{PPO}{short = PPO,
	long = proximal policy optimization}
\DeclareAcronym{REINFORCE}{short = REINFORCE,
	long = {\emph{RE}ward Increment $=$ \emph{N}onnegative \emph{F}actor $\times$ \emph{O}ffset \emph{R}einforcement $\times$ \emph{C}haracteristic \emph{E}ligibility}}
\DeclareAcronym{RTO}{short = RTO,
	long = real-time optimization,
	short-indefinite = an}	
\DeclareAcronym{SAC}{short = SAC,
	long = soft actor-critic}	
\DeclareAcronym{TD3}{short = TD3,
	long = twin-delayed DDPG}	
\DeclareAcronym{HER}{short = HER,
	long = hindsight experience replay}	
\DeclareAcronym{DPC}{short = DPC,
	long = differentiable predictive control}

\DeclareAcronym{GP}{short = GP,
	long = Gaussian process,
	long-plural-form = Gaussian processes}
\DeclareAcronym{RBF}{short = RBF,
	long = radial basis function,
	short-indefinite = an}	
\DeclareAcronym{SAE}{short = SAE,
	long = sparse autoencoder}	
\DeclareAcronym{DBN}{short = DBN,
	long = deep belief network}	
\DeclareAcronym{LSTM}{short = LSTM,
	long = long short-term memory}	
\DeclareAcronym{KL}{short = KL,
	long = Kullback-Leibler}
\DeclareAcronym{MDP}{short = MDP,
	long = Markov decision process,
	long-plural-form = Markov decision processes,
	short-indefinite = an}
\DeclareAcronym{pomdp}{short = POMDP,
	long = partially observed Markov decision process,
	long-plural-form = partially observed Markov decision processes,
	short-indefinite = a}
\DeclareAcronym{LQR}{short = LQR, 
	long = linear quadratic regulator}
\DeclareAcronym{LQ}{short = LQ, 
	long = linear quadratic}
\DeclareAcronym{DARE}{short = DARE, 
	long = discrete algebraic Riccati equation}
\DeclareAcronym{LTI}{short = LTI, 
	long = linear time-invariant,
	short-indefinite = an}
\DeclareAcronym{GPS}{short = GPS,
	long = guided policy search}	
\DeclareAcronym{GRU}{short = GRU,
	long = gated recurrent unit}	
\DeclareAcronym{ESN}{short = ESN,
	long = echo state network}	
\DeclareAcronym{ENN}{short = ENN,
	long = Elman neural network}	
	
\DeclareAcronym{CSTR}{short = CSTR,
	long = continuous stirred tank reactor}

\begin{document}
\begin{frontmatter}

\title{Why Goal-Conditioned Reinforcement Learning Works: Relation to Dual Control\thanksref{footnoteinfo}} 
%\title{Reinforcement learning with intrinsic rewards\thanksref{footnoteinfo}} 

\thanks[footnoteinfo]{This material is based upon work supported by the U.S. Department of Energy, Office of Science, Office of Fusion Energy Sciences under award number DE-SC0024472.}

\author[First]{Nathan P. Lawrence} 
\author[First]{Ali Mesbah} 

\address[First]{Department of Chemical and Biomolecular Engineering, University of California, Berkeley, CA 94720 USA (e-mail: \{nplawrence,mesbah\}@berkeley.edu).}

\begin{abstract}                % Abstract of 50--100 words
Goal-conditioned reinforcement learning (RL) concerns the problem of training an agent to maximize the probability of reaching target goal states.
This paper presents an analysis of the goal-conditioned setting based on optimal control.
In particular, we derive an optimality gap between more classical, often quadratic, objectives and the goal-conditioned reward, elucidating the success of goal-conditioned RL and why classical ``dense'' rewards can falter.
We then consider the partially observed Markov decision setting and connect state estimation to our probabilistic reward, making the goal-conditioned reward well suited to dual control problems.
The advantages of goal-conditioned policies are validated on nonlinear and uncertain environments using both RL and predictive control techniques.
\end{abstract}

\end{frontmatter}
%===============================================================================

\section{Introduction}

%% There are a number of predefined theorem-like environments in
%% ifacconf.cls:
%%
%% \begin{thm} ... \end{thm}            % Theorem
%% \begin{lem} ... \end{lem}            % Lemma
%% \begin{claim} ... \end{claim}        % Claim
%% \begin{conj} ... \end{conj}          % Conjecture
%% \begin{cor} ... \end{cor}            % Corollary
%% \begin{fact} ... \end{fact}          % Fact
%% \begin{hypo} ... \end{hypo}          % Hypothesis
%% \begin{prop} ... \end{prop}          % Proposition
%% \begin{crit} ... \end{crit}          % Criterion

% \begin{table}[hb]
% \begin{center}
% \caption{Margin settings}\label{tb:margins}
% \begin{tabular}{cccc}
% Page & Top & Bottom & Left/Right \\\hline
% First & 3.5 & 2.5 & 1.5 \\
% Rest & 2.5 & 2.5 & 1.5 \\ \hline
% \end{tabular}
% \end{center}
% \end{table}

%Optimal decision-making under uncertainty concerns numerous disciplines from \ac{RL} to dual control.
Dynamic programming is at the heart of optimal decision-making under uncertainty,
giving general optimality conditions for an agent to satisfy \citep{bertsekas2012dynamic}.
Traditionally, the optimal control problem consists of two distinct objects: a reward and an uncertain model.
% given a cost or reward function and uncertain dynamic model .
This paper sits in the general setting of dynamic programming, but reexamines the role of the reward function.
In particular, we leverage the dynamic uncertainty to estimate not only the hidden state, but also the probability of reaching target states.
% estimate probabilities of reaching target states.
% In particular, we design an ``intrinsic'' reward directly from the system uncertainty.
This results in
%rather than viewing the reward as a design variable separate from the system uncertainty, we use the dynamic uncertainty itself to formulate a more ``intrinsic'' reward.
a dual control problem in which the objective is an ``intrinsic'' goal-reaching probability that directly stems from the system uncertainty.

% and the system uncertainty are essentially the same object.

The notion of a probabilistic reward function has been proposed in the goal-conditioned \ac{RL}  literature \citep{eysenbach2022contrastive}, and has connections to indicator-type rewards \citep{eysenbach2021clearning, liu2022GoalConditionedReinforcement, lawrence2025view}.
However, solution methods by \cite{eysenbach2021clearning, eysenbach2022contrastive} are \emph{indirect}, employing contrastive learning techniques for learning policies.
This is because direct approaches may be intractable either due to a lack of model knowledge or the sparsity of the probabilistic reward function.
Meanwhile, classic control typically uses a quadratic cost to formulate optimal control problems; in the \ac{RL} literature these are so-called ``dense'' rewards.
While such rewards are not sparse, they can still be difficult to configure towards good performance, making them nuisance parameters in goal-reaching tasks \citep{forbes2015model}.

%This paper aims to give a clear explanation for why probabilistic, goal-oriented rewards work from the vantage point of classic optimal control.
Despite strong evidence that goal-oriented rewards are effective for solving complex control tasks, it is unclear why this is the case, or more precisely, why there is a significant performance gap between dense and probabilistic rewards.
This paper sheds light on probabilistic rewards from the vantage point of classic optimal control. 
We take a direct, value-based approach towards analyzing the goal-oriented objective.
More specifically, our contributions are summarized as follows:
\begin{enumerate}
	\item An inequality that lower bounds the objective of goal-oriented policies in terms of ``classical'' formulations, corroborating folklore that dense rewards may not be suitable for goal-oriented planning (\cref{sec:3.1}).
	\item A partially observed control formulation in which we draw connections between goal-oriented control and dual control. In this setting, the reward itself is intimately tied to state estimation, yielding a control problem defined entirely by system uncertainty (\cref{sec:dual}).
	\item Case studies in challenging nonlinear and uncertain environments (\cref{sec:methods}), emphasizing that the above benefits are related to the control objective itself and not necessarily the underlying algorithms (\cref{sec:results}). Similarly, we show how the same methods but with classical objective functions can falter.
\end{enumerate}

\subsection{A motivating example}
\label{subsec:motivation}

The crux of this paper is captured in the distinction between the following two control objectives:
\begin{equation}
	\exp \left( - \frac{1}{2} \sum_{t=0}^{\infty} \gamma^t \norm{x_t}^2 \right) \quad \text{vs.}\quad \sum_{t=0}^{\infty} \gamma^t \exp\left( - \frac{1}{2} \norm{x_t}^2 \right).
\end{equation}
Objectives of the former style are referred to as \emph{classic control}.
The inner term is a familiar quadratic cost, while the exponentiation relates the costs to probabilities, leading to maximum-entropy \citep{levine2018ReinforcementLearning}, path integral \citep{williams2017information}, dual \citep{todorov2008Generalduality},
 and robust \citep{jacobson1973Optimalstochastica} control formulations.

We refer to the latter objective as \emph{goal-oriented}, motivated as follows.
% The goal-oriented objective is motivated as follows.
An idealized reward for goal-directed behavior would be of an indicator type, providing positive feedback only when a goal state is achieved.
However, such feedback is vanishingly sparse, as in a continuous and stochastic environment a learning agent will never reach the precise goal (with probability one).
Instead, 
% due to the intrinsic uncertainty of the environment, 
the goal-oriented objective can be seen as a sum of (unnormalized) probability densities evaluated at the origin:
\begin{equation}
	e^{-\frac{1}{2\epsilon^2}\norm{x_t}^2} \approx \ind_{\{ \norm{x} < \epsilon\}} (x_t) = 
	\begin{cases}
	1 & \text{Goal is achieved}\\
	0 & \text{Otherwise}	
	\end{cases}
\label{eq:goalreward}
\end{equation}
where $\epsilon > 0$ is made arbitrarily small.
Indicator type objectives are common in RL \citep{liu2022GoalConditionedReinforcement}, usually treating the tolerance $\epsilon$ as a hyperparameter.

% Instead, this paper focuses on a goal-oriented framework with the density structure outlined above, as it reflects the intrinsic uncertainty of the environment, meaning it is the ``true'' objective, while the indicator type objective is an idealized representation.

To see the difference between the classical and goal-oriented objectives, suppose the initial state is far from the origin, but the agent reaches its goal in $T$ time steps. 
Then, the classic control objective becomes degenerate, as $\exp(-\frac{1}{2} \norm{x_0}^2)\exp(-\frac{1}{2} \gamma \norm{x_1}^2)\cdots \approx 0$, depending entirely on the initial state.
In contrast, the goal-oriented objective is approximately $\nicefrac{\gamma^T}{1-\gamma}$, depending entirely on $T$.
This simple example indicates that the classic control objective---which uses a so-called ``dense'' reward---is sparse in the initial state space, whereas the goal-oriented objective is sparse in the time horizon.
In this paper, we analyze these two objectives, showing that their respective performance is connected, but not equivalent.
In particular, we derive a lower bound to the goal-oriented objective in terms of the classic control objective, indicating that they are equivalent locally around the goal, but that the goal-oriented objective will be more suitable globally.
Moreover, we make a connection between the intrinsic uncertainty of the environment and the goal-oriented objective, resulting in a dual control problem.
% we leverage the intrinsic uncertainty of the environment to make precise connections between the goal-oriented objective and dual control.
% , meaning it is the ``true'' objective, while the indicator type objective is an idealized representation.

%The above scenario is treated in the context of \iac{pomdp}.

%In this simple example we see that the classic control objective is suitable in a local sense around the origin, whereas the goal-oriented objective is agnostic to its starting state 

\section{Problem setting}
% problem setting, discounted density over goal

We consider the problem of an agent interacting with an environment with the goal of maximizing the long-term probability of reaching a desired state.
The environment is represented as an uncertain dynamical system with states $x$, actions $u$, and observations $y$ that evolves according to the following transition and measurement functions:
\begin{equation}
\begin{split}
    x' &= f(x,u,\omega)\\
    y &= g(x,\nu),  
\label{eq:dynamics}
\end{split}
\end{equation}
where $\omega, \nu$ denote process and measurement uncertainty, respectively.
%Equivalently, we often refer to \cref{eq:dynamics} in terms of probability densities, writing $x' \sim \pp{p}{x'}{x,u}$ and $y \sim \pp{p}{y}{x}$.  
The agent only has access to past inputs and outputs, a collection known as the \emph{information state}:
\begin{equation}
	\mathcal{I}_t = \left[ y_0, \ldots, y_t, u_0, \ldots, u_{t-1} \right]
\end{equation}
with $\mathcal{I}_0 = \left[y_0\right]$.
The information state is impractical for decision-making, as it grows with time.
Instead, the information state can be used to characterize a \emph{belief state} (or \emph{hyperstate}), the conditional density over states:
\begin{equation}
	b_t = \pp{p}{x_t}{\mathcal{I}_t}.
\label{eq:hyperstate}
\end{equation}
That is, $b_t$ describes the uncertainty around the internal state $x_t$ given the current observed information $\mathcal{I}_t$.
The belief state can be computed through recursive Bayesian estimation as follows \citep{chen2003bayesian, mesbah2018Stochasticmodel}:
\begin{equation}
\begin{split}
	\pp{p}{x_t}{\mathcal{I}_t} &= \frac{\pp{p}{y_t}{x_t}\pp{p}{x_t}{\mathcal{I}_{t-1}, u_{t-1}}}{\pp{p}{y_t}{\mathcal{I}_{t-1}, u_{t-1}}} \\
	\pp{p}{x_{t+1}}{\mathcal{I}_{t}, u_t} &= \int \pp{p}{x_{t+1}}{x_t, u_t} \pp{p}{x_t}{\mathcal{I}_{t}} d x_t
\label{eq:bayes_est}
\end{split}
\end{equation}
with $\pp{p}{x_0}{\mathcal{I}_{-1}} = p(x_0)$.
It then follows that the belief state evolves recursively with time as actions are applied to the environment and new measurements are made available: $\{ b_0, u_0, y_1, b_1, u_1, y_2, \ldots \}$.
We write
\[
b_{t+1} = \mathcal{H}(b_t, u_t, y_{t+1})
\]
to represent this transition between belief states over time.

Based on the current belief state, and a slight abuse of notation, the agent generates an action from a policy $u \sim \pp{\pi}{u}{b}$.
From this action, a density over next states $\pp{p}{x'}{b, u}$ is available based on the transition model in \cref{eq:dynamics}, equivalent to \cref{eq:bayes_est}.
In particular, the agent aims to bring the unobserved state $x'$ to a goal state, namely, the origin $x' = 0$; we make this simplification without loss of generality, as we could reorient nonzero goals $x_g$ around the origin with a change of variables $\hat{x} = x - x_g$.
Naturally, based on \cref{eq:bayes_est}, the cost associated with the state-action tuple $(b, u)$ is the next-state density evaluated at the goal: 
\begin{equation}
    r(b, u) = \EE_{x \sim \pp{p}{x}{b}}\left[\pp{p}{x'=0}{x,u} \right].
\end{equation}
This leads to the goal-oriented control objective:
\begin{equation}
\begin{aligned}
    &\text{maximize} && J(\pi) = \mathbb{E}_{\pi}\left[ \sum_{t=0}^{\infty} \gamma^{t} \pp{p}{x_{t+1} = 0}{x_t,u_t} \right]\\
    &\text{where} &&  x_0 \sim p\left(x_0\right) \\
    & && u_t \sim \pp{\pi}{u_t}{b_t}\\ 
    & && b_{t+1} = \mathcal{H}(b_{t}, u_t, y_{t+1}),\\
\end{aligned}
\label{eq:mdp_objective}
\end{equation}
%% TODO after I finish examples, might be useful to reevaluate whether we want to evaluate stage cost at the current state or next state prediction
%\Cref{eq:mdp_objective} can be thought of as maximizing the probability of reaching the origin.
%Indeed, the objective \cref{eq:mdp_objective} can be rewritten as:
%\begin{equation}
%	J(\pi) = \log\left((1-\gamma)\sum_{t=0}^{\infty} \gamma^{t} \mathbb{E}_{\pi}\left[ \pp{p}{x_{t+1} = 0}{b_t,u_t} \right]\right),
%\end{equation}
%where the stage cost can be thought of as the probability density over states visited by the policy at a particular time step.
%The $(1 - \gamma)$ term then ensures the objective integrates   to $1$.
where the expectation is with respect to the policy and all system uncertainty in \cref{eq:dynamics}.
\Cref{eq:mdp_objective} is \iac{pomdp} \citep{krishnamurthy2025PartiallyObserved, lim2023optimality}, \iacs{MDP} over belief states rather than full state observations.
% In light of \cref{subsec:motivation}, \cref{eq:mdp_objective} is a goal-oriented objective, whereas a classic control setup would use as the stage cost $\log(\pp{p}{x_{t+1}}{x_t,u_t})$.
Next, we relate \cref{eq:mdp_objective} to classic control objectives, showing they are not equivalent.
%While the natural logarithm is increasing, it is applied at each stage, resulting in a different objective than \cref{eq:mdp_objective}.
%We expand on these points in the next section, explaining \cref{eq:mdp_objective} in relation to more familiar objectives.

\section{Analyzing the goal-oriented objective}
\label{sec:objective}
% relationship to classical control, relationship to dual control

\subsection{Connection to classical control}
\label{sec:3.1}

For brevity, this section assumes full (but uncertain) state measurements (with $\nu=0)$, that is, $b = x = y$.
The following inequalities and discussion are general and can be modified in light of the full uncertain system \cref{eq:dynamics}.
The following theorem establishes that \cref{eq:mdp_objective} is distinct from the classical objective that sums log-probabilities; this result will be a useful vehicle for reasoning about the optimal policies under differing objectives.

\begin{theorem}
    Assume the state transition density $p$ in \cref{eq:dynamics} is bounded. For any policy $\pi$ and $\gamma \in (0,1)$, the following inequality holds:
\begin{equation}
\begin{multlined}
(1 - \gamma) \EE_\pi \left[ \sum_{t = 0}^{\infty} \gamma^t \log \left( \pp{p}{x_{t+1} = 0}{x_t,u_t} \right) \right]\\
    \leq \log\left( (1-\gamma) \EE_\pi \left[ \sum_{t=0}^{\infty} \gamma^t \pp{p}{x_{t+1} = 0}{x_t,u_t} \right] \right) .
\end{multlined}
\label{eq:prob_bound}
\end{equation}
\label{thm:prob_bound}
\end{theorem}
\begin{pf}
    Inequality \cref{eq:prob_bound} follows by Jensen's inequality. Define $\rho_T^{-1} = \nicefrac{1-\gamma}{1 - \gamma^{T+1}}$. For $T = 1, 2, \ldots$ and any state-action trajectory $\{ x_0, u_0, x_1, u_1, \ldots\}$ under policy $\pi$, we have
\begin{equation}
\begin{multlined}    
    -\log\left( \rho_T^{-1} \sum_{t=0}^{T-1} \gamma^t \pp{p}{x_{t+1} = 0}{x_t,u_t} \right) \\
    \leq -\rho_T^{-1} \sum_{t=0}^{T-1} \gamma^t \log \left( \pp{p}{x_{t+1} = 0}{x_t,u_t} \right).
\end{multlined}
\end{equation}
Taking the limit as $T \to \infty$ and evaluating the expectation over trajectories, we apply Jensen's inequality once more to obtain the result.
\qed
\end{pf}
\begin{remark}
    $\gamma \in (0,1)$ plays an important role, ensuring the above sums are convex combinations of terms involving the transition density. In that vein, the result also applies to the average cost formulation.
\end{remark}
\begin{remark}
    Equality only holds in \cref{eq:prob_bound} over constant trajectories. This means the terms in \cref{eq:prob_bound} are equivalent locally around the goal, but result in different global control laws.
\end{remark}

Both sides of the inequality in \cref{eq:prob_bound} represent measures of concentration of the system $f$ around the origin by policy $\pi$.
The right-hand side views the entire control horizon as a probability density at the goal state $x=0$, while the left-hand side considers the stage-wise log-probability of reaching the goal state. 
As such, \cref{thm:prob_bound} provides a mechanism for evaluating policies under different objectives, and consequently, analyzing the respective optimal policies under each objective.
If the log-probability objective is tractable with finite cost, then so is the goal-oriented objective, meaning classic control can be a useful primitive for formulating goal-oriented policies.
We elucidate this with the next corollaries, relating \cref{eq:prob_bound} to the classical \ac{LQR} setting.

\begin{corollary}[Nonlinear goal-oriented policy]
    Under the assumptions of \cref{thm:prob_bound} and for a linear system $x_{t+1} = A x_t + B u_t + \omega_t$ with $\omega_t$ unit normal, the optimal goal-oriented policy achieves strictly better performance than any \ac{LQR} controller.
\label[corollary]{thm:nonlinear}
\end{corollary}

\begin{pf}
Assume the Bellman optimality equation is satisfied (see \cref{sec:dual}):
\begin{equation}
\max_{u \in \mathcal{U}} Q^{\star} (x,u) = \max_{u \in \mathcal{U}} \left\{ r(x,u) + \gamma \EE_{x'\sim\pp{p}{x'}{x,u}} \left[ V^{\star} (x') \right] \right\}.
\label{eq:bellman_nonlinear}
\end{equation}
Solving for the optimal action, we find:
\begin{align}
&0 = - \exp\left( -\frac{1}{2} \norm{A x + B u}^2 \right) \left( B^{\top} A x + B^{\top} B u \right) \\
&+ \gamma \EE_{x'\sim\pp{p}{x'}{x,u}} \left[ B^{\top} \grad V^{\star}(x') - \left( B^{\top} A x + B^{\top} B u \right) V^{\star} (x') \right].
\end{align}
This is a \emph{nonlinear} equation in $u$.
Therefore, in light of \cref{eq:prob_bound}, any \ac{LQR} controller is suboptimal with respect to the goal-oriented objective.
\qed
\end{pf}

\Cref{thm:nonlinear} establishes two key insights.
First, for a goal-oriented policy to be successful, the control horizon must be sufficiently long for the agent to ``see'' the goal.
This is because the action sensitivities contain the exponential term, which will remain degenerate until an action $T$ steps into the future brings the planned state near the goal.
Second, \cref{thm:prob_bound,thm:nonlinear} suggest a simple reward design strategy for goal-oriented learning.
For deterministic problems, the probability density over next states evaluated at the goal is intractable; instead, we can define the reward with a Gaussian-shaped structure and treat the covariance as a parameter, rather than an intrinsic quantity.
With this adjustment, we still retain the previous results with the following modified objectives.

\begin{corollary}[Goal-oriented reward design]
    Consider the state transition dynamics $f$ in \cref{eq:dynamics}. For any $M, R > 0$ and $\gamma \in (0,1)$, we have the following inequality:
\begin{equation}
\begin{multlined}
    -\frac{(1 - \gamma)}{2} \EE_\pi \left[ \sum_{t=0}^{\infty} \gamma^t \left( x_t^{\top} M x_t + u_t^{\top} R u_t \right) \right] \\ \leq  \log\left( (1-\gamma) \EE_\pi \left[ \sum_{t=0}^{\infty} \gamma^t \exp\left( - \frac{1}{2} x_{t}^{\top} M x_{t} \right)\right] \right).
\end{multlined}
\label{eq:lqr_bound}
\end{equation}
\label[corollary]{thm:lqr}
\end{corollary}

% \section{Direct approach by dynamic programming}
\subsection{Connection to dual control}
\label{sec:dual}

This section outlines the theoretical framework for solving \cref{eq:mdp_objective}, drawing connections to dual control.
Dual control refers to the problem of designing a policy that balances control and probing \citep{mesbah2018Stochasticmodel}.
In principle, a dual controller probes the system to reduce uncertainty, which, in turn, should enable more effective control.
Building on \cref{sec:objective}, we argue that the goal-oriented reward is well suited for the dual setting.

The optimal control problem in \cref{eq:mdp_objective} can be approached through stochastic dynamic programming \citep{bertsekas2012dynamic}.
The optimal (belief) state value function $V^\star$ mirrors \cref{eq:mdp_objective} except it maximizes the cumulative future reward beginning at any initial belief state.
Consequently, the optimal value function results in a one-step recursion known as the Bellman optimality equation:
\begin{equation}
\begin{multlined}
	V^\star (b) = \max_{u \in \mathcal{U}} \{ \EE_{x \sim \pp{p}{x}{b}}\left[  \pp{p}{x' = 0}{x, u} \right]  \quad\quad\quad\\+ \gamma \EE_{y' \sim \pp{p}{y'}{b,u}} \left[  V^\star( b' ) \right] \}.
\end{multlined}
\label{eq:bellman}
\end{equation}
Note the immediate reward depends on the distribution of inferred states; additionally, $V^\star$ has knowledge of future belief states.
Therefore, optimal actions satisfying \cref{eq:bellman} affect both the current state and the future system uncertainty.
This is known as the \emph{dual effect} \citep{feldbaum1963Dualcontrol, bar-shalom1974Dualeffect}.

% However, the Bellman equation \cref{eq:bellman} is generally intractable, inspiring approximate approaches broadly categorized as either \emph{implicit} or \emph{explicit} methods \citep{mesbah2018Stochasticmodel}.
% Explicit solutions design a tractable surrogate objective in a heuristic fashion, often based on \ac{MPC}, aimed at emulating key features of the optimal solution $V^\star$ such as planning and identification.
% Implicit approaches keep \cref{eq:bellman} at the forefront of the policy design.
% We expand on this class of methods next.

To see why the goal-oriented reward may be beneficial for dual control, we first make a minor adjustment to the reward.
Namely, let the goal-oriented reward be conditioned on the next observation: $\pp{p}{x'=0}{b, u, y'}$, which, as before, can also be computed according to \cref{eq:bayes_est}.
Then \cref{eq:bellman} becomes
\begin{equation}
\begin{multlined}
	V^\star (b) = \max_{u \in \mathcal{U}} \{ \EE_{y' \sim \pp{p}{y'}{b,u}}\left[  \pp{p}{x' = 0}{b, u, y'} + \gamma   V^\star( b' ) \right] \}.
\end{multlined}
\label{eq:bellman_nextobs}
\end{equation}
By the definition of expectation, we find the immediate reward is
\begin{equation}
	r(b,u) = \int \pp{p}{x' = 0}{b, u, y'} \pp{p}{y'}{b,u} d y'.
\label{eq:control_vs_est}
\end{equation}
The first term in the integral is related to control, while the second term is related to probing.
Indeed, this reward forces the agent to balance the goal of bringing the estimated state to the origin against the competing objective of also ensuring the state is likely based on the current measurement.
In this way, control and probing are tied together through the intrinsic system uncertainties \cref{eq:dynamics}; any reward function could be substituted into \cref{eq:control_vs_est}, but this connection would be lost.

\section{Algorithmic framework for goal-oriented policies}
\label{sec:methods}
% specific approximate solution methods: RL, DPC

This section briefly outlines the algorithmic ingredients used in this work, the descriptions of which are intentionally terse, but accompanied with suitable references.
These specific methods are not necessarily integral to the preceding sections; rather, they represent a broad and effective set of direct tools for approaching the goal-oriented objective in \cref{eq:mdp_objective}.
We emphasize a value-based approach to this objective, meaning the ensuing overview specifically targets the Bellman equation in \cref{eq:bellman}.

\textbf{Reinforcement learning.}\quad
Value-based \ac{RL} directly targets the Bellman optimality equation through a series of approximations.
Very broadly, some key approximations are the use of sample data in place of the true (usually unknown) dynamics, the use of neural networks to represent the optimal policy or value function, and the use of these networks to bootstrap value estimates involving intractable expectations of long planning horizons.
Deep \ac{RL} algorithms over continuous state and action spaces generally stem from \cite{silver2014DeterministicPolicy}.
The basic idea is to train a ``critic'' neural network $Q_{\phi}$ to approximate the cumulative future reward of following a policy, or ``actor'', network $\mu_\theta$.
The actor network is updated via the maximization problem: 
\begin{equation}
\begin{aligned}
    &\text{maximize} && \EE_{x \sim \mathcal{D}}\left[ Q_{\phi}(x, u) \right]\\
    &\text{where} && u = \mu_{\theta}(x) \\
	& && Q_{\phi}(x, u) \approx \EE\left[ \sum_{t=0}^\infty \gamma^t r(x_t,u_t) \mid \substack{x_0 = x\\u_0 = u} \right].
\end{aligned}
\label{eq:dpc_objective}
\end{equation}
This cycle repeats until convergence, often using a Gaussian exploration policy of the form $\pp{\pi}{\cdot}{x} = \mathcal{N}(\mu_\theta (x), \Sigma)$.

\textbf{Particle filter.}\quad
The goal-oriented optimal control problem \cref{eq:mdp_objective} is presented in the \ac{pomdp} setting, meaning the agent needs to estimate the underlying state in \cref{eq:dynamics}.
Moreover, the reward itself in \cref{eq:control_vs_est} must be estimated.
Particle filters are a versatile option, which, as shown next, will integrate seamlessly with the \ac{RL} framework for representing the value function.
Particle filters combine Monte Carlo sampling with Bayesian inference to approximate the posterior density of the hidden state in \cref{eq:bayes_est} \citep{chen2003bayesian}.
The essence of the approach is to use a set of ``particles'' $\{ x^{(0)}, \ldots, x^{(p-1)}\}$ in tandem with the transition and measurement model \cref{eq:dynamics} to approximate the posterior $\pp{p}{x_{t}}{\mathcal{I}_{t}}$ in \cref{eq:bayes_est}.
Namely, when a new measurement $y_{t}$ is available, the particles are propagated through the transition model under the action $u_{t-1}$ and scored based on the likelihood of the observation under the measurement density.
This leads to a set of weights $\{ w^{(0)}, \ldots, w^{(p-1)}\}$, with which the following approximation holds:
\begin{equation}
	\int f(x_{t}) \pp{p}{x_t}{I_t} d x_t \approx \sum_{i=0}^{p-1} f(x^{(i)}) w^{(i)}.
\label{eq:pf}
\end{equation}
For instance, setting $f$ to be identity yields a state estimate.

\textbf{Belief state value function representation.}\quad
The Bellman equation is generally intractable in the \ac{MDP} setting, and the problem is only exacerbated when extended to \acp{pomdp}.
The belief value function takes probability densities over states as input, rather than states themselves.
Therefore, not only is solving \cref{eq:bellman} intractable because it is a dynamic programming problem, but it also has an intractable input to its value function.
Following \cref{eq:bellman}, we highlight that the reward is an expectation with respect to the belief state. 
Therefore, we can view the entire Bellman recursion as an expectation over hidden states $x_t \sim \pp{p}{x_t}{b_t}$.
From \cref{eq:pf}, this inspires the representation
\begin{equation}
Q_\phi (b, u) = \frac{1}{p}\sum_{i=0}^{p-1} \hat{Q}_\phi (x^{(i)}) \approx \EE_{x \sim \pp{p}{x}{b}}\left[ \hat{Q} (x, u) \right],	
\label{eq:particle_critic}
\end{equation}
where $\hat{Q}_\phi$ is a critic network whose input size is the state dimension and the empirical mean is computed over a set of (resampled) particles.
Alternatively, one could use the representation $Q_\phi (b) = Q_{\phi} (x^{(0)}, \ldots, x^{(p-1)})$ at the risk of blowing up the input dimension, as in \cite{bayard2010Implicitdual}.
It is shown by \cite{lim2023optimality} that particle representations are an efficient vehicle for applying \ac{MDP} solution methods to \acp{pomdp}, as computational complexity is linear in the number of particles and optimality of the particle belief \ac{MDP} preserves convergence in the original \ac{pomdp}.

\section{Case studies}
\label{sec:results}

We present two case studies.
The first one validates the discrepancy between goal-oriented and classical objectives discussed in \cref{sec:objective} for a complex planning problem with no uncertainty.
The second tackles the full objective \cref{eq:mdp_objective}, incorporating estimation into a nonlinear control problem.
Corresponding code is openly available.\footnote{https://github.com/NPLawrence/goal-conditioned-control}

\subsection{Classic control example}

The aim of this example is to illustrate the differences between a quadratic and Gaussian-shaped stage cost.
\Cref{thm:prob_bound} demonstrates that \emph{if} a classical optimal control problem is tractable and stable, then the same setup but with a goal-oriented stage cost will lead to bounded goal tracking.
This example shows that the converse is not the case, namely, that the goal-oriented objective may be solvable when the classic control set up is intractable.

% In this example, we employ  \ac{DPC} \citep{drgona2022Differentiablepredictive}, 

We apply the \ac{DPC} \citep{drgona2022Differentiablepredictive} approach to the double inverted pendulum.
\Ac{DPC} is a deep learning-based alternative to \ac{MPC} wherein the \ac{MPC} objective is treated as a loss function over a class of policies. 
The double inverted pendulum is known to be challenging for \ac{DPC} due to its chaotic dynamics, making it a suitable benchmark for differentiating between the goal-oriented and classical objectives.
We assume complete model knowledge without uncertainty.
The objective is to bring the system from rest to the upright balancing position.
That is, we want the agent to bring the angle of both links to zero, or, $\cos(\theta) = 1$, a component-wise criterion across both links.
The goal-oriented objective is formulated with a Gaussian-shaped stage cost: $\log\left( \frac{1}{75} \sum_{t = 0}^{74} \exp\left( -\frac{1}{2} \norm{1 - \cos(\theta)}^2 \right)\right)$ and the classical control objective uses a quadratic stage cost: $\exp\left( - \frac{1}{75} \sum_{t=0}^{74} \frac{1}{2} \norm{1 - \cos(\theta)}^2\right)$.
Additionally, we consider two optimizers, the default Adam optimizer and SOAP \citep{vyas2025SOAPImproving}, an efficient combination of Adam and Gauss-Newton methods.

\begin{figure}
 \begin{center}
 \includegraphics[width=8.4cm]{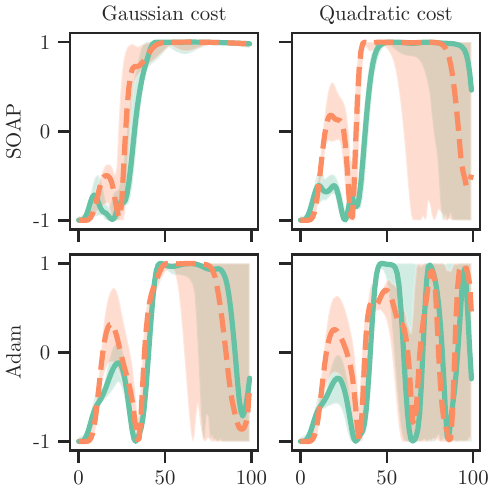}    % The printed column width is 8.4 cm.
 \caption{Time profiles of the double inverted pendulum across four policies. The y-axis is $\cos(\theta)$ for each link; $1 = \cos(0)$ is the upright position. The SOAP optimizer in tandem with a Gaussian stage cost performs the best, while the Adam optimizer with a quadratic stage cost is unable to stabilize the system.} 
 \label{fig:dip}
 \end{center}
 \end{figure}

As shown in \cref{fig:dip}, Adam alone is insufficient for both objectives, although the goal-oriented time profile shows fleeting signs of solving the task.
Similarly, the classic control objective in combination with SOAP can briefly stabilize the system.
Finally, the goal-oriented objective with SOAP leads to a policy that consistently stabilizes the system upright.
This indicates that the near-optimal goal-oriented policy is strictly better than the near-optimal classical policy, validating the discussion in \cref{sec:3.1}. 
We note, however, that these policies were not trained beyond $75$ time steps from the initial state, so they do not necessarily stabilize the system indefinitely.
In any case, the goal-oriented objective is shown to be effective at a highly nonlinear control task where a traditional objective proves to be intractable.
Future work may extend the control horizon by combining the goal-oriented policy with local stabilizing policies or \ac{RL} methods.

\subsection{Dual control example}

This example demonstrates the effectiveness of a probabilistic, goal-oriented reward in an uncertain environment.
We use a particle filter to approximate both the state and the reward using \cref{eq:pf}.
For the reward, we take $f$ to be an indicator-type function evaluated at each particle, essentially resulting in a histogram representation.
Next, we deploy the \ac{SAC} algorithm \citep{haarnoja2018Softactorcritic}, using the \ac{RL} state $b = \{ x^{(0)}, \ldots, x^{(p-1)}\}$ and the value representation in \cref{eq:particle_critic}.
Similarly, the actor is also an empirical mean over the particles.
Note the number of particles can be tuned due to this averaging strategy.

We consider \iac{CSTR} environment with $4$ states---concentrations $c_A$ and $c_B$, reactor temperature $T_R$, and coolant temperature $T_K$---and $2$ actions---normalized inflow $F$ and heat removed by coolant $\dot{Q}$; see \citep{lawrence2025view} or the GitHub repository for full details of the environment.
The objective is to achieve a desired concentration $c_B$.
The uncertainty is characterized by two multipliers $\alpha$ and $\beta$: $\alpha$ characterizes uncertainty in the activation energy for the reaction $A \to D$, while $\beta$ characterizes uncertainty in the rate coefficient for the reaction $A \to B$.
Formally, our model has a scenario-based structure with measurement noise: $\psi \sim p(\psi),\ x' = f_{\psi}(x,u),\ y = g(x) + \nu$, where $\psi = \{ \alpha, \beta\}$.

We train six agents. 
\textbf{Full:}\quad
For each training episode, the environment is randomly initialized with a set of fixed parameters $\psi \sim p(\psi)$.
Therefore, the agent must design actions such that the particles governed by the scenario-based model align with the unknown parameters, while also bringing the true concentration $c_B$ to a desired value. This is akin to the dual control problem.
\textbf{Partial:}\quad
The agent is trained only under the branching model above. 
That is, the ``true'' environment has time-varying parameters.
\textbf{Minimal:}\quad
The environment consists of a single parameter instance across all episodes and the agent has full state observations. This results in a certainty equivalence policy.
Further, for each experiment type above, we consider two rewards.
\textbf{Goal-conditioned:}\quad
\Cref{eq:control_vs_est} is written with the particle filter. In the ``minimal'' case, we use a Gaussian-shaped reward as in the previous example.
\textbf{Quadratic:}\quad
The same procedure is used, but with $f$ quadratic in \cref{eq:pf}.
All six agents are evaluated in the same way, namely, over $1000$ episodes each initialized with a random set of fixed system parameters $\alpha$ and $\beta$.

\begin{figure}
 \begin{center}
 \includegraphics[width=8.4cm]{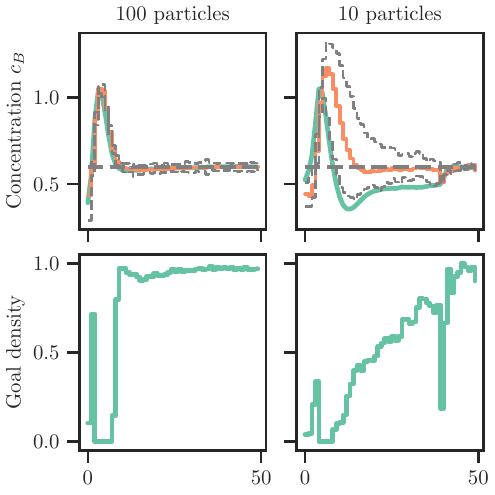}    % The printed column width is 8.4 cm.
 \caption{(Top) Time profiles of the same ``full'' agent using $100$ particles versus $10$ in the action selection process towards goal $c_B = 0.6$. The step-style lines show the mean state estimate inside the extreme (min \& max) particle values. The green (smooth) curve is the realized trajectory in the environment. (Bottom) Estimated reward over time.} 
 \label{fig:timeprofile_cstr}
 \end{center}
 \end{figure}

\begin{figure}
 \begin{center}
 \includegraphics[width=8.4cm]{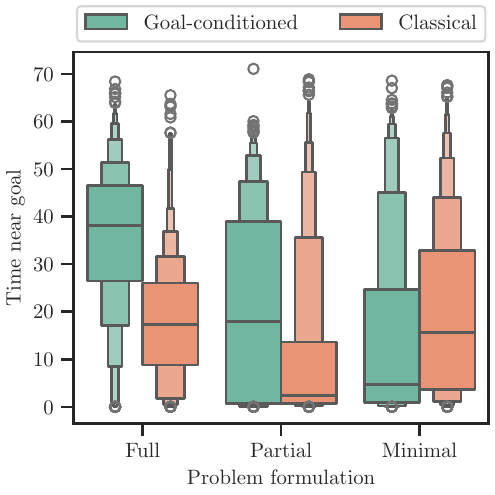}    % The printed column width is 8.4 cm.
 \caption{The distribution of time spent near goal across $6$ different agents,  each using $100$ particles.} 
 \label{fig:boxen_cstr}
 \end{center}
 \end{figure}

Fig. \ref{fig:timeprofile_cstr} shows the time profile of the ``full'' agent under the goal-conditioned reward. Both panels use the same actor, but evaluated over a different number of particles: $u = \sum_{i=0}^{p-1} \mu_\theta (x^{(i)})$ for $p = 10$ and $100$.
This shows the effectiveness and flexibility of ``democratizing'' action selection across particles.
The bottom panels show the (scaled) estimated reward based on the density \cref{eq:control_vs_est}.
Note the model trajectory under the $10$ particle case tracks the goal, but the reward is suboptimal until the true state aligns with the estimate, highlighting the tension between estimation and control in \cref{eq:control_vs_est}.
Next, \cref{fig:boxen_cstr} shows the distribution of time spent near the goal across all six agents, that is, the cumulative sum of $\exp\left(-\frac{1}{2\sigma^2}(0.6 - c_B)^2\right)$ over a trajectory where $c_B$ is the true concentration.
The main takeaway is that the goal-conditioned agents are able to improve in performance going from the minimal to the full case, indicating \cref{eq:mdp_objective} is indeed a higher fidelity objective towards goal-reaching policies versus their traditional quadratic counterparts. 
The classical quadratic variants tend to have similar performance across all agents.

\section{Conclusion}
This paper presented an analysis of goal-conditioned \ac{RL} in relation to optimal control, elucidating the performance gap between dense and probabilistic rewards.
Our results demonstrate the advantages of goal-oriented learning, both in a classical and dual control setting.
We adopted a value-based perspective of the goal-oriented problem.
This direct approach indicates that the advantages of the goal-oriented reward are intrinsic to the problem setting itself, rather than the specific algorithms used for policy design.

\begin{ack}
We thank Thomas Banker for helpful discussions.
\end{ack}

\bibliography{2025_ifac.bib}

\end{document}